\documentclass[12pt,a4paper,oneside]{article}
\usepackage{fullpage}
\usepackage[english]{babel} % English language/hyphenation.
\usepackage[T1]{fontenc} % Use 8-bit output encoding.
\usepackage[utf8]{inputenc} % Can use UTF-8 in the source files.
\usepackage[babel]{microtype} % Improves appearance of text.

\usepackage{url}
\usepackage{csquotes}
\usepackage{amsmath,amsthm,amssymb,bm,mathtools}
\usepackage[pdftex]{graphicx}
\graphicspath{{fig/}}
\usepackage{enumitem}
\usepackage{booktabs}
\usepackage{siunitx}

\usepackage{algorithm}
\usepackage{algpseudocode}

\usepackage{natbib}
\usepackage{hyperref}

\theoremstyle{plain}
\newtheorem{lemma}{Lemma}
\newtheorem{theorem}{Theorem}
\newtheorem{proposition}{Proposition}

\newcommand{\email}[1]{\href{mailto:#1}{\nolinkurl{#1}}}

\usepackage{xifthen}
\newcommand{\intint}[2]{\ensuremath{\{#1 \; .. \; #2 \}}}
\newcommand{\BigO}[2][]{\ensuremath{%
\ifthenelse{\isempty{#1}}{O( #2 )}{O_{ #1 }( #2 )}}}
\newcommand{\LittleO}[1]{\ensuremath{o(#1)}}
\newcommand{\BigOmega}[1]{\ensuremath{\Omega(#1)}}

\newcommand{\Set}[1]{\ensuremath{\mathbf{#1}}}
\newcommand{\Prob}[1]{\ensuremath{\mathbf{P}\left[ #1 \right]}}
\newcommand{\Exp}[2][]{\ensuremath{%
\ifthenelse{\isempty{#1}}{\mathbf{E}\left[ #2 \right]}{\mathbf{E}_{#1}\left[ #2 \right]}}}
\newcommand{\Cov}[2]{\ensuremath{\mathbf{Cov}\left[ #1, #2 \right]}}
\newcommand{\Var}[1]{\ensuremath{\mathbf{Var}\left[ #1 \right]}}
\newcommand{\Abs}[1]{\ensuremath{\lvert #1 \rvert}}
\newcommand{\Indic}[1]{\ensuremath{\mathbf{1} \left \{ #1 \right \} }}

\newcommand{\DExp}[1]{\ensuremath{\mathrm{Exp}( #1 )}}
\newcommand{\DGamma}[1]{\ensuremath{\mathrm{Gamma}( #1 )}}
\newcommand{\DBeta}[1]{\ensuremath{\mathrm{Beta}( #1 )}}
\newcommand{\DUniform}[1]{\ensuremath{\mathrm{U}( #1 )}}

\newcommand{\Disp}[2][]{\ensuremath{%
\ifthenelse{\isempty{#1}}{\Delta( #2 )}{\Delta_{#1}( #2 )}}}

\newcommand{\id}{\ensuremath{\mathrm{id}}}

\newcommand{\hsigma}{\ensuremath{\hat{\sigma}}}
\newcommand{\tsigma}{\ensuremath{\tilde{\sigma}}}

\DeclareMathOperator{\BT}{BT}
\DeclareMathOperator{\KL}{KL}
\DeclareMathOperator*{\Argmin}{arg\,min}
\DeclareMathOperator*{\Argmax}{arg\,max}
\DeclarePairedDelimiter{\ceil}{\lceil}{\rceil}

\newlist{enuminline}{enumerate*}{1}
\setlist[enuminline,1]{label=\itshape\alph*\upshape)}

\title{Just Sort It! A Simple and Effective Approach \\
to Active Preference Learning}
\author{
Lucas Maystre\thanks{School of Computer and Communication Sciences, EPFL, Switzerland.}\\
\email{lucas.maystre@epfl.ch}
\and
Matthias Grossglauser\footnotemark[1]\\
\email{matthias.grossglauser@epfl.ch}
}

\begin{document}
\maketitle

\begin{abstract} 
We address the problem of learning a ranking by using adaptively chosen pairwise comparisons.
Our goal is to recover the ranking accurately but to sample the comparisons sparingly.
If all comparison outcomes are consistent with the ranking, the optimal solution is to use an efficient sorting algorithm, such as Quicksort.
But how do sorting algorithms behave if some comparison outcomes are inconsistent with the ranking?
We give favorable guarantees for Quicksort for the popular Bradley--Terry model, under natural assumptions on the parameters.
Furthermore, we empirically demonstrate that sorting algorithms lead to a very simple and effective active learning strategy: repeatedly sort the items.
This strategy performs as well as state-of-the-art methods (and much better than random sampling) at a minuscule fraction of the computational cost.
\end{abstract} 

%%%%%%%%%%%%%%%%%%%%%%%%%%%%%%%%%%%%%%%%%%%%%%%%%%%%%%%%%%%%%%%%%%%%%%%%%
\section{Introduction}  %%%%%%%%%%%%%%%%%%%%%%%%%%%%%%%%%%%%%%%%%%%%%%%%%
\label{sec:intro}

The problem of recovering a ranking over $n$ items from noisy outcomes of pairwise comparisons has attracted, in the last century, much research interest, driven by applications in sports \citep{elo1978rating}, social sciences \citep{thurstone1927law, salganik2015wiki} and---more recently---recommender systems \citep{houlsby2012collaborative}.
Whereas pairwise comparison models and related inference algorithms have been extensively studied, the issue of \emph{which pairwise comparisons to sample}, also known as active learning, has received significantly less attention.
To understand the potential benefits of adaptively selecting samples, consider the case where comparison outcomes are noiseless, i.e., consistent with a linear order on a set of $n$ items.
If pairs of items are selected at random, it is necessary to collect \BigOmega{n^2} comparisons to recover the ranking \citep{alon1994linear}.
In contrast, by using an efficient sorting algorithm, \BigO{n \log n} adaptively chosen comparisons are sufficient.
In this work, we demonstrate that sorting algorithms can also be helpful in the \emph{noisy} setting, where some comparison outcomes are inconsistent with the ranking: despite errors, sorting algorithms tend to select informative samples.
We focus on the Bradley--Terry (BT) model, a widely-used probabilistic model of comparison outcomes.
In this model, each item is associated with a parameter on the real line, and the probability of observing an incorrect outcome decreases as the distance between the items' parameters increases.

First, we study the output of a single execution of Quicksort when comparison outcomes are generated from a BT model, under the assumption that the distance between adjacent parameters is (stochastically) uniform across the ranking.
We measure the quality of a ranking estimate by its displacement with respect to the ground truth, i.e., the sum of rank differences.
We show that Quicksort's output is a good approximation to the ground-truth ranking: no method comparing every pair of items at most once can do better (up to constant factors).
Furthermore, we show that by aggregating \BigO{\log^5 n} independent runs of Quicksort, it is possible to recover the exact rank for all but a vanishing fraction of the items.
These theoretical results suggest that adaptive sampling is able to bring a substantial acceleration to the learning process.

Second, we propose a practical active-learning (AL) strategy that consists of repeatedly sorting the items.
We evaluate our sorting-based method on three datasets and compare it to existing AL methods.
We observe that \emph{all} the strategies that we consider lead to better ranking estimates noticeably faster than random sampling.
However, most strategies are challenging to operate and computationally expensive, thus hindering wider adoption \citep{schein2007active}.
In this regard, sorting-based AL stands out, as
\begin{enuminline}
\item it is computationally-speaking as inexpensive as random sampling, 
\item it is trivial to implement, and
\item it requires no tuning of hyperparameters.
\end{enuminline}

\subsection{Preliminaries and Notation}

We consider $n$ items that are represented by consecutive integers $[n] = \{1, \ldots, n\}$.
Without loss of generality, we assume that the items are ranked by increasing preference\footnote{
This convention greatly simplifies the notation throughout the paper, but differs from that used in most of the preference learning literature.
In our paper, the item with rank $1$ is the \emph{worst}.}, i.e., $i < j$ means that $j$ is (in expectation) preferred to $i$.
When $j$ is preferred to $i$ as a result of a pairwise comparison, we denote the observation by $i \prec j$.
If $i < j$, we say that $i \prec j$ is a \emph{consistent} outcome and $j \prec i$ an \emph{inconsistent} (incorrect) outcome.
In most of the paper, pairwise comparison outcomes follow a Bradley--Terry model with parameters $\bm{\theta} = \begin{bmatrix} \theta_1 & \cdots & \theta_n \end{bmatrix} \in \Set{R}^n$, denoted $\BT(\bm{\theta})$.
The parameters $\theta_1 < \cdots < \theta_n$ represent the utilities of items $1, \ldots, n$, and the probability of observing the outcome $i \prec j$ is
\begin{align*}
p(i \prec j \mid \bm{\theta}) = \frac{1}{1 + \exp[-(\theta_j - \theta_i)]}.
\end{align*}
The probability of observing an inconsistent comparison decreases with the distance between the items.
This captures the intuitive notion that some pairs of items are easy to compare and some are more difficult \citep{zermelo1928berechnung, bradley1952rank}.

A ranking $\sigma$ is a function that maps an item to its rank, i.e., $\sigma(i) =$ rank of item $i$.
The (ground-truth) identity ranking is denoted by \id, i.e. $\id(i) = i$.
To measure the quality of a ranking $\sigma$ with respect to the ground-truth, we consider the \emph{displacement}
\begin{align*}
\Disp{\sigma} = \sum_{i=1}^n | \sigma(i) - i |,
\end{align*}
also known as Spearman's footrule distance.
Another metric widely used in practice is the Kendall--Tau distance, defined as
$K(\sigma) = \sum_{i < j} \Indic{\sigma(i) > \sigma(j)}$.
Both metrics are equivalent up to a factor of two\footnote{$\Disp{\sigma} / 2 \le K(\sigma) \le \Disp{\sigma}$ \citep{diaconis1977spearman}.}, such that bounds on \Disp{\sigma} also hold for $K(\sigma)$ up to constant factors.

Finally, we say that an event $A$ holds \emph{with high probability} if $\Prob{A} \to 1$ as $n \to \infty$.
For a random variable $X$ and a sequence of numbers $a_n$, we say that $X = \BigO{a_n}$ with high probability if $\Prob{\Abs{X} \le c a_n} \to 1$ as $n \to \infty$ for some constant $c$ that does not depend on $n$.

\paragraph{Outline of the paper.}
We begin by briefly reviewing related literature in Section~\ref{sec:relwork}.
Next, in Section~\ref{sec:theory}, we study the displacement of Quicksort's output under noisy comparisons.
In Section~\ref{sec:experiments}, we empirically evaluate several AL strategies on three datasets.
Finally, we conclude in Section~\ref{sec:conclusion}.

%%%%%%%%%%%%%%%%%%%%%%%%%%%%%%%%%%%%%%%%%%%%%%%%%%%%%%%%%%%%%%%%%%%%%%%%%
\section{Related Work}  %%%%%%%%%%%%%%%%%%%%%%%%%%%%%%%%%%%%%%%%%%%%%%%%%
\label{sec:relwork}

\paragraph{Passive setting.}
Recently, there have been a number of results on the sample complexity of the BT model, based on the assumption that all pairs of items are chosen \emph{before} any comparison outcome is revealed
\citep{negahban2012iterative, hajek2014minimax, rajkumar2014statistical, vojnovic2016parameter}.
In general, these results reveal that choosing pairs of items uniformly at random is essentially optimal.
Furthermore, they suggest that the ranking induced by the BT model cannot be recovered with less than \BigOmega{n^2} comparisons.
Our work shows that by \emph{adaptively} selecting pairs based on observed outcomes, we observe substantial gains.

\paragraph{Active preference learning.}
AL approaches for learning a ranking based on noisy comparison outcomes have been studied under various assumptions.
\citet{braverman2008noisy} examine a model where outcomes of pairwise comparisons are flipped with a small, constant probability.
\citet{ailon2012active} considers an adversarial setting (comparison outcomes can be arbitrary) and investigates AL in the context of finding a ranking that minimizes the number of inconsistent outcomes, also known as the minimum feedback-arc set problem on tournaments (MFAST).
These theoretical studies imply, in their respective settings, that \BigO{n \log^k n} comparison outcomes are enough to recover a near-optimal ranking.
%They design an algorithm that recovers the maximum-likelihood ranking in polynomial-time using only \BigO{n \log n} comparisons, but its running time is impractical.
\citet{jamieson2011active} propose an efficient active-ranking algorithm that is applicable if items can be embedded in $\Set{R}^d$ (e.g., using $d$ features) and assuming that admissible rankings satisfy some geometric constraints.
\citet{wang2014active} study a collaborative preference-learning problem and show that a variant of uncertainty sampling (a well-known AL strategy) works well for their problem.
In this work, we assume that we do not have access to item features and that comparison outcomes follow a single BT model.

\paragraph{Bayesian methods.}
From a practical standpoint, Bayesian methods provide an effective way to select informative samples \citep{mackay1992bayesian}.
However, they can be difficult to scale if the number of items is large.
Work on Bayesian active preference learning includes
\citet{chu2005extensions}, \citet{houlsby2012collaborative}, \citet{salimans2012collaborative} and \citet{chen2013pairwise}.
%While the methods in these papers generally assume that outcomes follow Thurstone's model \citep{thurstone1927law}, they are easy to extend to the BT model.
We compare our AL strategy to these methods in Section~\ref{sec:experiments}.

\paragraph{Multi-armed bandit.}
The \emph{dueling bandit} problem \citep{yue2009karmed} is somewhat related to our work.
In this problem, the goal is to identify the best item based on noisy comparison outcomes, using as few adaptively chosen samples as possible.
Two recent papers also extend the problem to that of recovering the entire ranking (instead of only the top element).
The work of \citet{szorenyi2015online} is the closest to ours, as it also uses the BT model.
One of their results is similar to our Theorem~\ref{thm:multidisp}: They show that a quasi-linear number of comparisons is sufficient to recover the true ranking, under some conditions on $\bm{\theta}$.
\citet{heckel2016active} investigate a non-parametric model and develop some theoretical guarantees.
In contrast to these works, our paper studies practical comparison budgets: we give theoretical guarantees for the output obtained from a single call to Quicksort, and in our experiments we never exceed $\approx 10$ calls.

\paragraph{Quicksort.}
The Quicksort algorithm \citep{hoare1962quicksort} is one of the most widely studied sorting procedures.
Quicksort has been shown to produce useful rankings beyond classic sorting problems.
For example, \citet{ailon2008aggregating} show that Quicksort produces (in expectation) a $3$-approximation to the MFAST problem.
Quicksort combined with BT comparison outcomes has also been proposed as a probabilistic ranking model \citep{ailon2008reconciling}.
We take advantage of some of the properties of this ranking model in order to derive the theoretical results of Section~\ref{sec:theory}.

%%%%%%%%%%%%%%%%%%%%%%%%%%%%%%%%%%%%%%%%%%%%%%%%%%%%%%%%%%%%%%%%%%%%%%%%%
\section{Theoretical Results}  %%%%%%%%%%%%%%%%%%%%%%%%%%%%%%%%%%%%%%%%%%
\label{sec:theory}

In this section, we begin by studying the behavior and output of Quicksort under inconsistent comparison outcomes, without any assumptions on the noise generating process.
Then, starting in Section~\ref{sec:poisson}, we focus on comparison outcomes generated by the BT model.
Most full proofs are deferred to Appendix~\ref{app:proofs}.

\algnewcommand{\IIf}[1]{\State\algorithmicif\ #1\ \algorithmicthen}
\begin{algorithm}[t]
   \caption{Quicksort}
   \label{alg:quicksort}
\begin{algorithmic}[1]
   \Require set of items $V$
   \IIf{$\lvert V \rvert < 2$} \Return list($V$)
   \Comment{Terminating case.}
   \State $L \gets \varnothing, R \gets \varnothing$
   \State $p \gets $ element of $V$ selected uniformly at random \label{line:pivot}
   \For{$i \in V \setminus \{ p \}$} \label{line:startpart}
     \If{$i \prec p$} \label{line:comp}
     \Comment{Pairwise comparison.}
       \State $L \gets L \cup \{i\}$
     \Else
       \State $R \gets R \cup \{i\}$
     \EndIf
   \EndFor  \label{line:stoppart}
   \State \Return $\text{Quicksort}(L) \cdot p \cdot \text{Quicksort}(R)$ \label{line:return}
\end{algorithmic}
\end{algorithm}

Quicksort (Algorithm~\ref{alg:quicksort}) is best described as a recursive procedure.
At each step of the recursion, a \emph{pivot} item $p$ is chosen uniformly at random (line \ref{line:pivot}).
Then, during the \emph{partition} operation (lines \ref{line:startpart}--\ref{line:stoppart}), every other item is compared to $p$ and added to the set $L$ or $R$, depending on the outcome.
If all comparison outcomes are consistent, it is well-known that Quicksort terminates after sampling \BigO{n \log n} comparisons with high probability.
What happens if we drop the consistency assumption?
The following two lemmas state that these key properties remain valid, no matter which (and how many) comparison outcomes are inconsistent.

\begin{lemma}
\label{lem:termination}
Quicksort always terminates and samples each of the $n(n\!-\!1) / 2$ possible comparisons at most once.
\end{lemma}

\begin{proof}
The proof is identical to the consistent setting.
Consider the state of $L$ and $R$ at the end of a partition operation.
Because $\Abs{L} + \Abs{R} = \Abs{V} - 1$, the recursive calls are made on sets of items of strictly decreasing cardinality, and the algorithm terminates after a finite number of steps.
Furthermore, suppose that Quicksort samples an outcome for the pair $(i, j)$.
Then either $i$ or $j$ is the pivot in a partition operation.
In either case, the pivot is not included in the recursive calls, which ensures that $(i, j)$ cannot be compared again.
\end{proof}

\begin{lemma}
\label{lem:samplecomp}
%If the comparison outcomes are independent of the query order,
Quicksort samples \BigO{n \log n} comparisons w.h.p.
\end{lemma}

\begin{proof}[Proof (sketch).]
We follow a standard analysis of Quicksort \citep[see, e.g.,][Section 3.3.3]{dubhashi2009concentration}.
With high probability, we choose a ``good'' pivot (i.e., one that results in a balanced partition) a constant fraction of the time.
In this case, the depth of the call tree is \BigO{\log n}.
As there are at most $n$ comparisons at each level of the call tree, we conclude that Quicksort uses \BigO{n \log n} comparisons in total.
With respect to the standard proof, we need some additional work to formalize the notion of ``good'' pivot to the setting where comparison outcomes are not consistent with a linear order.
\end{proof}

Lemma~\ref{lem:samplecomp} complements Theorem~$3$ in \citet{ailon2010preference}, which states that Quicksort samples \BigO{n \log n} in expectation.
These results might suggest that \emph{all} properties of Quicksort carry over to the noisy setting.
This is not the case.
For example, although Quicksort uses approximately $2n \ln n$ comparisons on average in the noiseless setting \citep{sedgewick2011algorithms}, this number can be distinctly different with inconsistent comparison outcomes\footnote{E.g., if comparison outcomes are uniformly random, all items are ``good'' pivots w.h.p., and the average number of comparisons will be closer to $n \log_2 n$ on average, for large $n$.}.

Quicksort (and efficient sorting algorithms in general) infer most pairs of items' relative position by transitivity and thus rely heavily on the consistency of comparison outcomes.
In the noisy case, it is therefore important to precisely understand the effect of an inconsistent outcome on the output of the algorithm; this effect extends beyond the pair of items whose comparison outcome was inconsistent.
For this purpose, the next Lemma bounds the displacement of Quicksort's output as a function of the inconsistent outcomes.

\begin{lemma}
\label{lem:dispbound}
Let $E$ be the set of pairs sampled by Quicksort and whose outcome is inconsistent with \id.
Let $\sigma$ be the output.
Then,
\begin{align*}
\Disp{\sigma} \le 2 \sum_{\mathclap{(i, j) \in E}}\; \Abs{i - j}
\end{align*}
\end{lemma}

\begin{proof}[Proof (sketch).]
Consider the first partition operation, with pivot $p$, resulting in partitions $L$ and $R$.
Denote the errors made during this partition operation by $E_1$.
We can show that the displacement is bounded by
\begin{align*}
\Disp{\sigma} \le \Disp[L]{\sigma} + \Disp[R]{\sigma} + 2 \;\sum_{\mathclap{(i, j) \in E_1}} \Abs{i - j},
\end{align*}
where \Disp[L]{\sigma} and \Disp[R]{\sigma} represent the displacement of the ordering induced by $\sigma$ on $L$ and $R$, respectively.
In other words, the total displacement can be decomposed into a term that represents the ``local'' displacement due to the partition operation and into two terms that account for errors in the recursive calls.
We obtain the desired result by recursively bounding \Disp[L]{\sigma} and \Disp[R]{\sigma}.
\end{proof}

Informally, Lemma~\ref{lem:dispbound} states that the displacement can be bounded by a sum of ``local shifts'' due to the inconsistent outcomes and that the price to pay for any information inferred by transitivity is bounded by a factor two.
Lemma~\ref{lem:dispbound} is a crucial component of our subsequent analysis of BT noise, and we believe that it can be useful in order to investigate Quicksort under a wide variety of other noise generating processes.

%%%%%%%%%%%%%%%%%%%%%%%%%%%%%%%%%%%%%%%%%%%%%%%%%%%%%%%%%%%%%%%%%%%%%%%%%
\subsection{Displacement in the Poisson Model}
\label{sec:poisson}

From here on, we assume that comparison outcomes are generated from $\BT(\bm{\theta})$.
Clearly, any results on the displacement of a ranking estimated from samples of a BT model will depend on $\bm{\theta}$; it is easy to construct a model instance for which it is arbitrarily hard to recover the ranking, by choosing parameters sufficiently close to each other.
Our approach is as follows.
We postulate a family of distributions over $\bm{\theta}$, and we give bounds on the displacement that hold with high probability.

We suppose that comparison outcomes are (in expectation) \emph{uniformly noisy across the ranking}: i.e., comparing two elements at the bottom is (a priori) as difficult as comparing two elements at the top or in the middle.
This means that the probability distribution over parameters $\theta_1, \ldots, \theta_n$ results in (random) distances \Abs{\theta_{i+k} - \theta_i} that depend only on $k$.
One such distribution arises if the parameters are drawn from a Poisson point process of rate $\lambda$.
That is,
\begin{align}
\label{eq:poisson}
\text{i.i.d.}\; x_1, \ldots, x_{n-1} \sim \DExp{\lambda}, \qquad
\theta_i = \sum_{k=1}^{i-1} x_k.
\end{align}
The average distance between two items separated by $k$ positions in the ordering is $\Exp{\theta_{i+k} - \theta_i} = k / \lambda$.
%More generally, the distance \Abs{\theta_{k+i} - \theta_i} has distribution $\DGamma{k, \lambda}$.
Although the distance between adjacent items is constant in expectation, we allow some parameters to be arbitrarily close\footnote{
In particular, the expected minimum distance between two items (i.e., the $\min$ of $n$ exponential r.v.s) decreases as $(n\lambda)^{-1}$ as $n$ increases.}.
The parameter $\lambda$ controls the expected level of noise; a large $\lambda$ is likely to result in a larger number of inconsistent outcomes.
Although the precise choice of this Poisson model is driven by tractability concerns, in Section~\ref{sec:iidunif} we argue that it is essentially equivalent to choosing the parameters independently and uniformly at random in the interval $[0, (n+1) / \lambda]$, when $\lambda$ is fixed and $n$ is large.
We are now ready to state our main result.

\begin{theorem}
\label{thm:quickdisp}
Let $\bm{\theta}$ be sampled from a Poisson point process of rate $\lambda$.
Let $\sigma$ be the output of Quicksort using comparison outcomes sampled from $\BT(\bm{\theta})$.
Then, w.h.p.,
\begin{align}
\Disp{\sigma}
    &= \BigO{ \lambda^2 n }, \label{eq:qdtot} \\
\max_{i} \; \Abs{\sigma(i) - i}
    &= \BigO{ \lambda \log n }. \label{eq:qdmax}
\end{align}
\end{theorem}

\begin{proof}[Proof (sketch)]
Let $z_{ij}$ be the indicator random variable of the event ``the comparison between $i$ and $j$ results in an error'', and let $d_{ij} = \Abs{\theta_i - \theta_j}$.
The distance $d_{ij}$ is a sum of \Abs{i - j} exponential random variables, i.e., $d_{ij} \sim \DGamma{\Abs{i - j}, \lambda}$, and we can show that
\begin{align*}
\Exp{z_{ij}} &= \Exp{\frac{1}{1 + \exp(d_{ij})}}
    \le \Exp{\exp(-d_{ij})} = (1 + 1/\lambda)^{-\Abs{i - j}}.
\end{align*}
Using Lemma~\ref{lem:dispbound} and the fact that every pair of items is compared at most once, we find
\begin{align*}
\Exp{\Delta}
    \le 2 \sum_{i < j}\; \Abs{i - j} \Exp{z_{ij}}
    \le 2n \sum_{k = 0}^{\infty} k (1 + 1/\lambda)^{-k} = 2n \lambda (\lambda + 1).
\end{align*}
The random variables $\{ z_{ij} \}$ are not unconditionally independent (they are independent when conditioned on $\bm{\theta}$) but, with some more work, we can show that $\Var{\Delta} = O(n)$.
By using a Chebyshev bound, \eqref{eq:qdtot} follows.

In order to prove \eqref{eq:qdmax}, we take advantage of a theorem due to \citet{ailon2008reconciling} which states that
\begin{align*}
\Prob{ \sigma(i) < \sigma(j) \mid \bm{\theta} } = p(i \prec j \mid \bm{\theta}),
\end{align*}
even if $i$ and $j$ were not directly compared with each other.
We use a Chernoff bound on $d_{ij}$ to show that the relative order between any two items separated by at least $O(\lambda \log n)$ positions is correct with high probability.
The second part of the claim follows easily.
\end{proof}

Note that any method that compares each pair of items at most once results in a ranking estimate $\tau$ with displacement $\Disp{\tau} = \BigOmega{n}$ with high probability: As there is only a single (possibly inconsistent) comparison outcome between each pair of adjacent items, it is likely that a constant fraction of the items will be ranked incorrectly, resulting in a displacement that grows linearly in $n$.
Hence, our bound on \Disp{\sigma} shows that Quicksort is order-optimal (in $n$).

In light of Theorem~\ref{thm:quickdisp}, a natural question to ask is as follows.
How many comparisons are needed in order to find the correct ranking?
Clearly, finding the exact ranking is difficult: in fact, \BigOmega{n} comparison outcomes are necessary to discriminate the closest pair of items reliably (see Appendix~\ref{app:adjacent}).
As such, we will focus on finding a ranking that matches the ground truth everywhere, except at a vanishing fraction of the items.

\begin{algorithm}[t]
   \caption{Multisort}
   \label{alg:multisort}
\begin{algorithmic}[1]
   \Require set of items $V$, number of iterations $m$
   \State $S \gets \varnothing$
   \For{$k = 1, \ldots, m$}
     \State $\sigma \gets \text{Quicksort}(V)$
     \State $S \gets S \cup \{ \sigma \}$
   \EndFor
   \State \Return Copeland aggregation of $S$
\end{algorithmic}
\end{algorithm}

Multiple runs of Quicksort likely produce different outputs, because of the noisy comparison outcomes and because the algorithm itself is randomized (the pivot selection is random).
By aggregating $m$ independent outputs of Quicksort, is it possible to produce a better ranking estimate?
Similarly to \citet{szorenyi2015online}, we combine the $m$ outputs  $\sigma_1, \ldots, \sigma_m$ into an aggregate ranking $\hsigma$ using Copeland's method.
The method assigns, to each item, a score that corresponds to the number of items that it beats in a majority of the rankings, and it then ranks the items by increasing score \citep{copeland1951reasonable}.
We call the procedure Multisort and describe it in Algorithm~\ref{alg:multisort}.

\begin{theorem}
\label{thm:multidisp}
Let $\bm{\theta}$ be sampled from a Poisson point process of rate $\lambda$.
Let $\hsigma$ be the output of Multisort using $m = \BigO{\lambda^2 \log^5 n}$ and comparison outcomes sampled from $\BT(\bm{\theta})$.
Then, w.h.p.,
\begin{align*}
\Disp{\hsigma} = \LittleO{\lambda n}.
\end{align*}
\end{theorem}

\begin{proof}[Proof (sketch)]
We use results on the order statistics of the distances $x_1, \ldots, x_{n-1}$ between successive items, as defined in \eqref{eq:poisson}, to partition the items into two disjoint subsets $B$ and $G$.
The set $B$ contains a vanishing $(1/\log^2 n)$-fraction of ``bad'' items that are difficult to order.
The set $G$ is such that the smallest distance $d_{ij}$ from any item $i \in G$ to any other item $j \in [n]$ is bounded from below by $c / (\lambda \log^2 n)$.
We can show that with $m = \BigO{\lambda^2 \log^5 n}$, for any $i \in G$ and $j \in [n]$ we have $i < j \iff \sigma(i) < \sigma(j)$ in a majority of the Quicksort outputs (with high probability).
This implies that $\hsigma(i) = i$ for all $i \in G$ with high probability.
Using \eqref{eq:qdmax} for items in $B$, we have
\begin{align*}
\Disp{\hsigma} = \Abs{B} \cdot \BigO{\lambda \log n} = \BigO{\lambda n / \log n}
\end{align*}
with high probability.
\end{proof}

Theorem~\ref{thm:multidisp} states that all but a vanishing fraction of items are correctly ranked using \BigO{\lambda^2 n \log^6 n} comparisons.
This result should be compared to the \BigOmega{n^2} comparisons needed if samples are selected uniformly at random.

\paragraph{Empirical validation.}

\begin{figure*}[t]
\centering
\includegraphics[width=\linewidth]{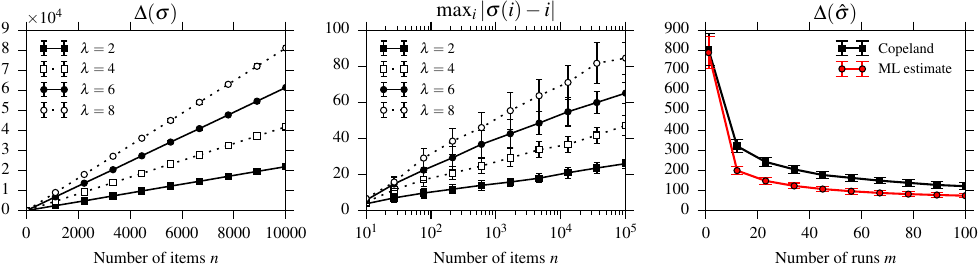}
\caption{
Empirical validation of Theorem~\ref{thm:quickdisp} and illustration of Theorem~\ref{thm:multidisp}.
Every simulation is repeated $50$ times, and we report the mean and the standard deviation.
Left and middle: total and maximum displacement (respectively) for increasing $n$ and different values of $\lambda$.
Right: displacement of the aggregate ranking $\hsigma$ for increasing $m$, fixing $n = \num{200}$ and $\lambda = \num{4}$ and using two different aggregation rules.
}
\label{fig:bounds}
\end{figure*}

In Figure~\ref{fig:bounds}, we illustrate the results of Theorems~\ref{thm:quickdisp} and~\ref{thm:multidisp} by running simulations for increasing $n$ and different values of $\lambda$.
The bound on \Disp{\sigma} is tight in $n$, but the dependence on $\lambda$ appears to be linear rather than quadratic.
The bound on $\max_i \Abs{\sigma(i) - i}$ appears to be tight in $n$ and $\lambda$.
Finally, we compare the Copeland aggregation of $m$ outputs of Quicksort with the ranking induced by the maximum-likelihood (ML) estimate, inferred from the outcomes of all the pairwise comparisons sampled by the $m$ runs.
Although the ranking induced by the ML estimate does not benefit from the guarantees of Theorem~\ref{thm:multidisp}, it performs better in practice.
We will make use of this observation in Section~\ref{sec:experiments}.

%%%%%%%%%%%%%%%%%%%%%%%%%%%%%%%%%%%%%%%%%%%%%%%%%%%%%%%%%%%%%%%%%%%%%%%%%
\subsection{Independent Uniformly-Distributed Parameters}
\label{sec:iidunif}

A different (perhaps more natural) assumption on the parameters $\bm{\theta}$ is to consider that they are drawn independently and uniformly at random over some interval.
That is,
\begin{align*}
\text{i.i.d.}\; \bar{\theta}_1, \ldots, \bar{\theta}_n \sim \DUniform{0, (n+1) / \lambda},
\end{align*}
with $\theta_1, \ldots, \theta_n$ the order statistics of $\bar{\bm{\theta}}$, i.e., the random variables arranged in increasing order.
From some elementary results on the joint distribution of order statistics \citep[see, e.g.,][]{arnold2008first}, we see that
\begin{align*}
\Abs{\theta_{i+k} - \theta_{i}} \sim (n+1) / \lambda \cdot \DBeta{k, n - k + 1},
\end{align*}
i.e., a Beta random variable rescaled between $0$ and $(n+1) / \lambda$.
Letting $f_{k,n}(x)$ be the probability density of $\Abs{\theta_{i+k} - \theta_{i}}$, we have, for any fixed $k$ and $\lambda$,
\begin{align*}
f_{k,n}(x) \propto x^{k-1} \left[ 1 - \frac{\lambda x}{n + 1} \right]^{n - k} \xrightarrow{n \to \infty} x^{k-1} e^{-\lambda x}.
\end{align*}
We recognize the functional form of the density of a \DGamma{k, \lambda} distribution.
Hence, the Poisson model and the i.i.d. uniform model are essentially equivalent for fixed $\lambda$ and large $n$, and we can expect the results developed in Section~\ref{sec:poisson} to hold under this distribution as well.

%%%%%%%%%%%%%%%%%%%%%%%%%%%%%%%%%%%%%%%%%%%%%%%%%%%%%%%%%%%%%%%%%%%%%%%%%
\section{Experimental Results}  %%%%%%%%%%%%%%%%%%%%%%%%%%%%%%%%%%%%%%%%%
\label{sec:experiments}

In practice, the comparison budget for estimating a ranking from noisy data might typically be larger than that for a single call to Quicksort, and it might not exactly match the number of comparisons required to run a given number of calls to Quicksort to completion.
Building upon the observations made at the end of Section~\ref{sec:poisson}, we suggest the following practical active-learning strategy:
for a budget of $c$ pairwise comparisons, run the sorting procedure repeatedly until the budget is depleted (the last call might have to be truncated).
Then, retain only the set of $c$ comparison pairs and their outcomes and discard the rankings produced by the sorting procedure.
The final ranking estimate is then induced from the ML estimate over the set of $c$ comparison outcomes.

In this section, we demonstrate the effectiveness of this sampling strategy on synthetic and real-world data.
In particular, we show that it is comparable to existing AL strategies at a minuscule fraction of the computational cost.

%%%%%%%%%%%%%%%%%%%%%%%%%%%%%%%%%%%%%%%%%%%%%%%%%%%%%%%%%%%%%%%%%%%%%%%%%
\subsection{Competing Sampling Strategies}

To assess the relative merits of our sorting-based strategy, we consider three strategies that we believe are representative of the state of the art in active preference learning.

\paragraph{Uncertainty sampling.}
Developed in the context of classification tasks, this popular active-learning heuristic suggests to greedily sample the point that lies closest to the decision boundary \citep{settles2012active}.
In the context of a ranking task, this corresponds to sampling the pair of items whose relative order is most uncertain.
After $t$ observations, given an estimate of model parameters $\bm{\theta}^t$, the strategy selects the $(t\!+\!1)$-st pair uniformly at random in
\begin{align*}
\Argmin_{i \ne j} \Abs{\theta^t_i - \theta^t_j}.
\end{align*}
This set can be computed in time \BigO{n \log n} by sorting the parameters.
The parameters themselves need to be estimated, e.g., using (penalized) ML inference that in practice can be the dominating cost.

\paragraph{Bayesian methods.}
If we have access to a full posterior distribution $q^t(\bm{\theta})$ instead of a point estimate $\bm{\theta}^t$, we can take advantage of the extra information on the uncertainty of the parameters to improve the selection strategy.
A principled approach to AL consists of sampling the point that maximizes the expected information gain \citep{mackay1992bayesian}.
That is, the pair of items at iteration $t+1$ is selected in
\begin{align}
\label{eq:entreduc}
\Argmax_{i \ne j} H(q^t) - \Exp{H(q^{t+1})},
\end{align}
where $H(\cdot)$ denotes the entropy function.
A conceptually similar but slightly different selection strategy is given by \citet{chen2013pairwise}.
Letting $q_{ij}$ be the marginal distribution of $(\theta_i, \theta_j)$, the pair is selected in
\begin{align}
\label{eq:kldiv}
\Argmax_{i \ne j} \Exp{ \KL(q^{t+1}_{ij} \Vert q^t_{ij}) },
\end{align}
where $\KL(\cdot)$ denotes the Kullback--Leibler divergence.
Computing the exact posterior is not analytically tractable for the BT model, but a Gaussian approximation can be found in time \BigO{n^3}.
Criteria \eqref{eq:entreduc} and \eqref{eq:kldiv} can be computed in constant time for each pair of items.
The dominating cost is again that of estimating $\bm{\theta}$ (or, in this case, $q(\bm{\theta})$).

In addition to these existing AL strategies, we also include in our experiments a variation of our sorting-based strategy that uses Mergesort instead of Quicksort.
In the noiseless setting, Mergesort is known to use on average $\approx \num{39}~\%$ fewer comparisons than Quicksort per run \citep{knuth1998art}, but it does not benefit from the theoretical guarantees developed in Section~\ref{sec:theory}.

%%%%%%%%%%%%%%%%%%%%%%%%%%%%%%%%%%%%%%%%%%%%%%%%%%%%%%%%%%%%%%%%%%%%%%%%%
\subsection{Running Time}

In this section, we briefly discuss the running time of the methods.
We implement ML and Bayesian approximate inference algorithms for the BT model as a Python library\footnote{See: \url{http://lucas.maystre.ch/choix}.}.
For ML inference, we find that the fastest running time is achieved by a truncated Newton algorithm (even for large $n$).
For approximate Bayesian inference, we use a variant of the expectation-propagation algorithm outlined by \citet{chu2005extensions}.
All experiments are performed on a server with a \num{12}-core Xeon X5670 processor running at \num{2.93}~GHz.
Numerical computations take advantage of the Intel Math Kernel Library.

We illustrate the running time of AL strategies as follows.
For $n \in \{10^2, 10^3, 10^4 \}$, we generate outcomes for $n$ comparisons pairs chosen uniformly at random among $n$ items.
For each strategy, we then measure the time it takes to select the $(n\!+\!1)$-st pair of items adaptively.
The results are presented in Table~\ref{tab:runningtime}.
Note that these numbers are intended to be considered as orders of magnitude, rather than exact values, as they depend on the particular combination of software and hardware that we use.
The running time of the Bayesian AL strategies exceed \num{10} hours for $n = 10^4$ and the calls were stopped ahead of completion.
Our sorting-based methods, like random sampling, are the only AL strategies whose running time is constant for increasing $n$ (and for increasing $c$).
In fact, their running time is negligible in comparison to the other strategies, including uncertainty sampling.

\begin{table}[t]
  \caption{
Time (in seconds) to select the $(n\!+\!1)$-st pair.
Values indicated by $\varepsilon$ are below $10^{-5}$.
See text for details.
}
  \vspace{2mm}
  \label{tab:runningtime}
  \centering
  \begin{tabular}{l ccc}
    \toprule
                  & \multicolumn{3}{c}{$T$ [s]} \\
                    \cmidrule(l){2-4}
    Strategy      & $n = 10^2$ & $n = 10^3$ & $n = 10^4$ \\
    \midrule
    uncertainty   & \num{0.05}      & \num{0.5}        & \num{11}      \\
    entropy       & \num{0.3}       & \num{40}         & ---           \\
    KL-divergence & \num{0.9}       & \num{71}         & ---           \\
    Mergesort     & $\varepsilon$   & $\varepsilon$    & $\varepsilon$ \\
    Quicksort     & $\varepsilon$   & $\varepsilon$    & $\varepsilon$ \\
    random        & $\varepsilon$   & $\varepsilon$    & $\varepsilon$ \\
    \bottomrule
  \end{tabular}
\end{table}

%%%%%%%%%%%%%%%%%%%%%%%%%%%%%%%%%%%%%%%%%%%%%%%%%%%%%%%%%%%%%%%%%%%%%%%%%
\subsection{Empirical Evaluation}

We now investigate three datasets and measure the displacement of rankings estimated from adaptively-chosen samples, as a function of the budget $c$.
Note that in order to use uncertainty sampling and Bayesian methods, it is necessary to choose a regularization strength or prior variance in the inference step.
Different values can result in drastically different outcomes (in particular for uncertainty sampling) and, in practice, choosing a good value can be a significant challenge\footnote{Observe that our sorting-based approach is entirely parameter-free and is therefore not affected by this issue.}.
In the following, we report results for the values that worked best \emph{a posteriori}.

\paragraph{Synthetic dataset.}

\begin{figure}[t]
\centering
\includegraphics[width=0.67\linewidth]{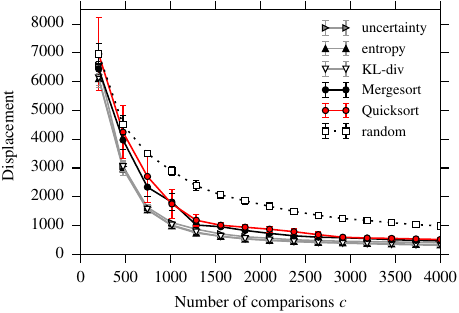}
\caption{
Synthetic dataset with $\lambda = 5$ and $n = 200$.
The experiment is repeated \num{10} times, and we report the mean and the standard deviation.
Compared to random sampling, AL results in significantly better rankings for a given budget $c$.
}
\label{fig:baselines}
\end{figure}

We generate $n$ i.i.d. parameters $\theta_1, \ldots, \theta_n$ uniformly in $[0, (n+1) / \lambda]$ and draw samples from $\BT(\bm{\theta})$.
The ground-truth ranking is the one induced by the parameters.
Figure~\ref{fig:baselines} presents results for $n = \num{200}$ and $\lambda = \num{5}$ (plots for different values of $\lambda$ are presented in Appendix~\ref{app:figures}, and are qualitatively similar).
In comparison to random sampling, AL is very effective and results in significantly better ranking estimates for any given number of comparisons.
The two Bayesian methods, though being the most computationally expensive, perform the best for all values of $c$, but are nearly indistinguishable from uncertainty sampling.
The two sorting-based strategies perform similarly (with a small edge for Mergesort).
They are slightly worse than the Bayesian methods but are still able to reap most of the benefits of active learning.

\paragraph{Sushi dataset.}

\begin{figure*}[t]
\centering
\includegraphics[width=\linewidth]{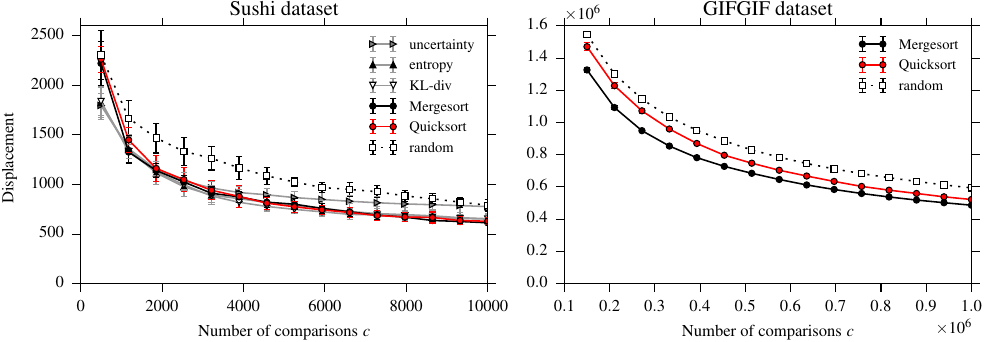}
\caption{
Results on two real-world datasets.
Every experiment is repeated \num{10} times, and we report the mean and the standard deviation.
Left: on the sushi dataset, sorting-based and Bayesian AL strategies have near-identical performance starting from $c \approx \num{1000}$.
Right: on the GIFGIF dataset, most AL strategies are computationally too expensive---except for sorting-based methods.
}
\label{fig:datasets}
\end{figure*}

Next, we consider a dataset of Sushi preferences \citep{kamishima2009efficient}.
In this dataset, \num{5000} respondents give a strict ordering over \num{10} different types of sushi.
These \num{10} sushi are chosen among a larger set of $n = \num{100}$ items.
To suit our purposes, we decompose each $10$-way partial ranking into pairwise comparisons, resulting in \num{225000} comparison outcomes.
We use all comparisons to fit a BT model that induces a ground-truth ranking\footnote{
The BT-induced ranking is almost the same as that obtained using the Copeland score.
The results are very similar if the Copeland aggregation is used as ground truth.}.

The comparisons are dense, and there is at least one comparison outcome for almost all pairs.
When an outcome for pair $(i,j)$ is requested, we sample uniformly at random over all outcomes observed for this pair.
In the rare case where no outcome is available, we return $i \prec j$ with probability $1/2$.
This enables us to compare sampling strategies in a realistic setting, where the assumptions of the BT model do not necessarily hold anymore.

Results are shown in Figure~\ref{fig:datasets} (left).
Once again, active learning performs noticeably better than random sampling.
On this real-world dataset, the performance of our sorting-based strategies is indistinguishable from that of the Bayesian methods, after completing one entire call to the sorting procedure (slightly less than \num{1000} comparisons).
This result should be interpreted in light of the time needed to select all $10^4$ pairs: a fraction of a second for sorting-based strategies, and several hours for the Bayesian methods.
Finally, we observe that the performance of uncertainty sampling progressively degrades as $c$ increases.
A detailed analysis reveals that uncertainty sampling increasingly focuses on a small set of hard-to-discriminate pairs, symptomatic of a well-known issue \citep{settles2012active}.

\paragraph{GIFGIF dataset.}

GIFGIF\footnote{See \url{http://www.gif.gf/}.
Data available at \url{http://lucas.maystre.ch/gifgif-data}.} is a project of the MIT Media Lab that aims at explaining the emotions communicated by a collection of animated GIF images.
Users of the website are shown a prompt with two images and a question, ``Which better expresses $x$?'' where $x$ is one of 17 emotions.
The users can click on either image, or use a third option, \emph{neither}.
To date, over three million comparison outcomes have been collected.
For the purpose of our experiment, we restrict ourselves to a single emotion, \emph{happiness}; and we ignore outcomes that resulted in \emph{neither}.
We consider \num{106887} comparison outcomes over $n = \num{6120}$ items---a significant increase in scale compared to the Sushi dataset.

As the data, despite a relatively large number of comparisons, remains sparse (less than 20 comparisons per item on average), we proceed as follows.
We fit a BT model by using all the available comparisons and use the induced ranking as ground truth.
We then generate new, synthetic comparison outcomes from the BT model.
In this sense, the experiment enables us to compare sampling strategies by using a large BT model with realistic parameters.
The large number of items makes uncertainty sampling and the two Bayesian methods prohibitively expensive.
We try a simplified, computationally less expensive version of uncertainty sampling where, at every iteration, each item is compared to its two closest neighbors, but this heuristic fails spectacularly: The resulting displacement is over $5\times$ larger than random sampling for $c = 10^6$, and is therefore not reported here (see Appendix~\ref{app:figures}).

Figure~\ref{fig:datasets} (right) compares the displacement of random sampling to that of the two sorting-based sampling strategies for increasing $c$.
The adaptive sampling approaches perform systematically better.
After $10^6$ comparisons, the displacement of random sampling is \num{14}~\% and \num{23}~\% larger than that of Quicksort and Mergesort, respectively.
Conversely, in order to reach any target displacement, Mergesort requires approximately $2 \times$ fewer comparisons than random sampling.

%%%%%%%%%%%%%%%%%%%%%%%%%%%%%%%%%%%%%%%%%%%%%%%%%%%%%%%%%%%%%%%%%%%%%%%%%
\section{Conclusion}  %%%%%%%%%%%%%%%%%%%%%%%%%%%%%%%%%%%%%%%%%%%%%%%%%%%
\label{sec:conclusion}

In this work, we demonstrate that active learning can substantively speed up the task of learning a ranking from noisy comparisons gains---both in theory and in practice.
With the advent of large-scale crowdsourced ranking surveys, exemplified by GIFGIF and wiki surveys \citep{salganik2015wiki}, there is a clear need for practical AL strategies.
However, existing methods are complex and computationally expensive to operate even for a reasonable number of items (a few thousands).
We show that a deceptively simple idea---repeatedly sorting the items---is able to bring in all the benefits of active learning, is trivial to implement, and is computationally no more expensive that random sampling.
Therefore, we believe that our method can be broadly useful for machine-learning practitioners interested in ranking problems.

\paragraph{Acknowledgments.}
We thank Holly Cogliati-Bauereis, Ksenia Konyushkova, Brunella Spinelli and anonymous reviewers for careful proofreading and helpful comments.

\appendix
%%%%%%%%%%%%%%%%%%%%%%%%%%%%%%%%%%%%%%%%%%%%%%%%%%%%%%%%%%%%%%%%%%%%%%%%%
\section{Proofs}  %%%%%%%%%%%%%%%%%%%%%%%%%%%%%%%%%%%%%%%%%%%%%%%%%%%%%%%
\label{app:proofs}

Section~\ref{sec:pflemmas} contains the proofs of Lemmas~\ref{lem:samplecomp} and~\ref{lem:dispbound}.
Section~\ref{sec:pfqdisp} presents the proof for our result on the displacement of the output of a single call to Quicksort (Theorem~\ref{thm:quickdisp}), and Section~\ref{sec:pfmdisp} that of our result on the displacement of the Copeland aggregation of multiple outputs.

%%%%%%%%%%%%%%%%%%%%%%%%%%%%%%%%%%%%%%%%%%%%%%%%%%%%%%%%%%%%%%%%%%%%%%%%%
\subsection{Lemmas~\ref{lem:samplecomp} and~\ref{lem:dispbound}}
\label{sec:pflemmas}

We start by briefly presenting a result from graph theory that will be useful in the proof of Lemma~\ref{lem:samplecomp}.
A \emph{tournament} is a directed graph obtained by assigning a direction to every edge of a complete graph.
The \emph{score sequence} of a tournament is defined as the nondecreasing sequence of the vertices' outdegrees.
The following proposition is due to \citet{landau1953dominance}.

\begin{proposition}
\label{prop:landau}
Let $(s_1, \ldots, s_n)$ with $0 \le s_1 \le \cdots \le s_n$ be the score sequence of a tournament on $n$ vertices.
Then,
\begin{align*}
\frac{k - 1}{2} \le s_k \le \frac{n + k - 2}{2} \quad \forall\ k \in [n].
\end{align*}
\end{proposition}

We use a tournament on $n$ vertices to represent the outcome of a comparison between each pair of items.
In particular, we represent the outcome $i \prec j$ by an edge $(i, j)$.
In this case, the outdegree of a vertex $i$ corresponds to the number of items which ``won'' in a comparison against $i$.
Note that the comparison outcomes do not need to be transitive, i.e., the tournament can contain cycles.

The proof of Lemma~\ref{lem:samplecomp} is adapted from standard results on Quicksort, see, e.g., \citet[][Section 3.3.3]{dubhashi2009concentration}.
These results are based on the fact that it is likely that the random choice of pivot leads to a well-balanced partition into subsets $L$ and $R$.
In our setting, the comparison outcomes do not need to be consistent with an ordering of the items, therefore we cannot use the standard argument based on the pivot's \emph{rank}.
Instead, we use the tournament representation of the comparison outcomes and analyze the pivot's \emph{out-degree} (using Proposition~\ref{prop:landau}) to ensure that the partition is balanced often enough.

\begin{proof}[Proof of Lemma~\ref{lem:samplecomp}]
We show that the maximum call depth of Quicksort is at most $\ceil{48 \log n}$ with high probability.
The statement follows by noticing that at most $n$ comparisons are used at each level of the call tree.

By Lemma~\ref{lem:termination}, Quicksort samples a comparison outcome for each pair of items at most once.
Therefore, we can represent these (a priori unobserved) pairwise outcomes as a tournament $T = ([n], A)$.
At each step of the recursion, we select a pivot $p$ uniformly at random in the set $V$ (line~\ref{line:pivot}), and compare it to the rest of the items in the set (line~\ref{line:comp}).
Let $T_V$ denote the subgraph of $T$ induced by $V$.
Given that the comparison outcomes follow from the edges of the tournament, $L$ is equal to the set of incoming neighbors of $p$ in $T_V$.
(Correspondingly, $R$ is equal to the set of the outgoing neighbors.)
Hence, the outdegree of $p$ in $T_V$ determines how balanced the partition is.
The probability that the outdegree of $p$ lies in the middle half of the score sequence is $1/2$, and if it does, Proposition~\ref{prop:landau} tells us that
\begin{align*}
\frac{\Abs{V} - 7}{8} \le \mathrm{outdeg}(p) \le \frac{7 \Abs{V} - 5}{8}.
\end{align*}
In this case, at the end of the partition \Abs{L} and \Abs{R} are of size at most $7\Abs{V}/8$, and in at most $\log_{8/7}(n) \le 8 \log n $ such partitions we get to a subset of size one and match the terminating case.
Even though we do not select the pivot in the middle half every time, it is unlikely that more than $c \cdot 8 \log n$ recursions are needed (for some small constant $c$) to select the pivot in the middle range at least $8 \log n$ times.
Let $z_d$ i.i.d $\sim \text{Bern}(1/2)$ be the indicator variable for the event ``the pivot is selected in the middle half at level of recursion $d$''.
Using a Chernoff bound, we have
\begin{align*}
\Prob{\sum_{d=1}^{\ceil{48 \log n}} z_d \le 8 \log n} \le \frac{1}{n^2},
\end{align*}
i.e., the depth of a leaf in the call tree is at most $\ceil{48 \log n}$ with probability at least $1 - 1/n^2$.
As there are at most $n$ leaves in the tree, the \emph{maximum} depth is bounded by the same value with probability at least $1 - 1/n$.
\end{proof}

In order to prove Lemma~\ref{lem:dispbound}, we introduce some additional notation.
For any $\sigma \in \Set{S}_n$ and $V \subseteq [n]$, let $\sigma_V : V \to \{1, \ldots, \Abs{V}\}$ be the ordering induced by $\sigma$ on $V$.
We generalize the definition of displacement as
\begin{align*}
\Disp[V]{\sigma, \tau} = \sum_{i \in V} \Abs{\sigma_V(i) - \tau_V(i)}.
\end{align*}
For conciseness, we use the shorthand $\Disp[V]{\sigma} \doteq \Disp[V]{\sigma, \id}$, where $\id$ is the identity permutation.

\begin{proof}[Proof of Lemma~\ref{lem:dispbound}]
Denote by $\mathcal{V}$ the collection of working sets that were used as input to one of the recursive calls to Quicksort.
For $V \in \mathcal{V}$, let $E_V$ be the set of pairs sampled by Quicksort to partition $V$ and which results in an error.
Note that $E_V \cap E_{V'} = \varnothing$ for $V \ne V'$, and that $\bigcup_V E_V = E$.
We will show that for all $V \in \mathcal{V}$,
\begin{align}
\label{eq:recursive}
\Disp[V]{\sigma} \le \Disp[L]{\sigma} + \Disp[R]{\sigma} + 2 \sum_{\mathclap{(i, j) \in E_V}}\; \Abs{i - j},
\end{align}
where $L$ and $R$ are the two sets obtained at the end of the partition operation.
The lemma follows by taking $V = [n]$ and recursively bounding \Disp[L]{\sigma} and \Disp[R]{\sigma}.

Consider the partition operation on $V$, with pivot $p$, resulting in partitions $L$ and $R$.
Let $\tsigma$ be the ordering on $V$ that
\begin{enuminline}
\item \label{itm:tsa} ranks $L$ at the bottom, $p$ in the middle and $R$ at the top, and
\item \label{itm:tsb} matches the identity permutation on $L$ and $R$, i.e., $\Disp[L]{\tsigma} = \Disp[R]{\tsigma} = 0$.
\end{enuminline}
In a sense, $\tsigma$ is the ordering that would be obtained if there were no further errors in the remaining recursive calls.
Using the triangle inequality, we have that
\begin{align}
\label{eq:tridisp}
\Disp[V]{\sigma} \le \Disp[V]{\sigma, \tsigma} + \Disp[V]{\tsigma}.
\end{align}
By definition of $\tsigma$, we have that
\begin{align}
\begin{split}
\label{eq:recdisp}
\Disp[V]{\sigma, \tsigma}
    = \Disp[L]{\sigma, \tsigma} + \Disp[R]{\sigma, \tsigma}
    = \Disp[L]{\sigma} + \Disp[R]{\sigma},
\end{split}
\end{align}
where the first equality follows from \ref{itm:tsa}, and the second follows from \ref{itm:tsb}.

\begin{figure}[t]
\centering
\includegraphics{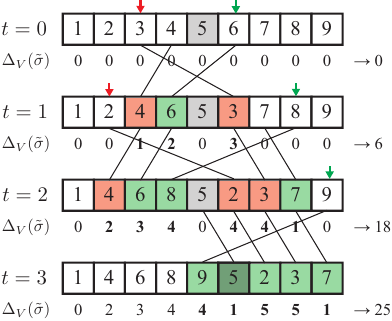}
\caption{Illustration of the decomposition of \Disp[V]{\tsigma} into contributions of individual errors over a sequence of steps.
In this example, $V = \{ 1, \ldots, 9 \}$, $p = 5$ and there are five errors.
At step $t = 1$, we process the errors $(5, 3)$ and $(5, 6)$;
at step $t = 2$, we process the errors $(5, 2)$ and $(5, 8)$, and finally, at step $t = 3$, we process the error $(5, 9)$.
The shifts caused by an error are highlighted in red and green.
In this example, $\Disp[V]{\tsigma} = 25 < 2 \sum_{(i,j) \in E_V} \Abs{i - j} = 26$.
}
\label{fig:lemma3}
\end{figure}

Finally, we bound $\Disp[V]{\tsigma}$.
Let $E_V^- = \{ (p, i) \in E_V : i < p \}$, and similarly $E_V^+ = \{ (p, i) \in E_V : i > p \}$.
Without loss of generality, we can assume that $V$ consists of consecutive integers, and that $\kappa \doteq \Abs{E_V^-} \le \Abs{E_V^+}$.
We proceed as follows: starting from the ranking $\id_V$, we progressively incorporate errors into the ranking, ending with $\tsigma$ once all errors have been treated.
To understand the impact of each error on \Disp[V]{\tsigma}, we look at errors in the following specific sequence.
\begin{enumerate}
\item At steps $t = 1, \ldots, \kappa$, we consider the $t$-th ``smallest'' errors in $E_V^-$ and $E_V^+$.
That is, we process $(p, i) \in E_V^-$ and $(p, i') \in E_V^+$ such that $\Abs{p - i}$ and $\Abs{p - i'}$, respectively, are smallest among errors not yet treated.

\item At steps $t = \kappa + 1, \ldots, \Abs{E_V^+}$, we process the remaining errors in $E_V^+$, once again in increasing order of distance to  $p$.
\end{enumerate}
Figure~\ref{fig:lemma3} illustrates the state of the ranking at different steps on a concrete example.
We start with the first case, i.e., $t \le \kappa$.
The effect of the errors $(p, i)$ and $(p, i')$ on $\Disp[V]{\tsigma}$ is as follows.
\begin{itemize}
\item All items $j < i$ and $j > i'$ are not affected by the two errors: their position remains the same.

\item The position of the pivot $p$ remains the same, as the two errors balance out.

\item Item $i$ is shifted by $\Abs{p - i} + 1$ positions to the right, just right of $p$.
Similary, item $i'$ is shifted by $\Abs{p - i'} + 1$ positions to the left, just left of $p$.

\item The $\Abs{p - i} - 1$ items that are between $p$ (excluded) and $i$ are shifted by $1$ position to the left.
Similarly, the $\Abs{p - i'} - 1$ items that are between $p$ and $i'$ are shifted by $1$ position to the right.
\end{itemize}
Hence, the two errors contribute $2 ( \Abs{p - i} + \Abs{p - i'} )$ towards $\Disp[V]{\tsigma}$.
Now consider the second case, when $t > \kappa$.
The effect of an error $(p, i)$ is as follows.
\begin{itemize}
\item All items $j > i$ and all the items on the left of $p$ are not affected by the error: their position remains the same.

\item The (at most) $\Abs{p - i}$ items that are between $p$ (included) and $i$ are shifted by $1$ position to the right.

\item Item $i$ is shifted by at most $\Abs{p - i}$ positions to the left, just left of $p$.
\end{itemize}
As a result, the error contributes at most $2 \Abs{p - i}$ to the displacement.
Adding up the contributions of all the errors, it follows that
\begin{align}
\label{eq:localdisp}
\Disp[V]{\tsigma} \le 2 \sum_{\mathclap{(i,j) \in E_V}}\; \Abs{i - j}.
\end{align}
Combining \eqref{eq:recdisp} and \eqref{eq:localdisp} using \eqref{eq:tridisp} we obtain \eqref{eq:recursive}, which concludes the proof.
\end{proof}

%%%%%%%%%%%%%%%%%%%%%%%%%%%%%%%%%%%%%%%%%%%%%%%%%%%%%%%%%%%%%%%%%%%%%%%%%
\subsection{Theorem \ref{thm:quickdisp}}
\label{sec:pfqdisp}

From now on, we focus on parameters drawn from a Poisson process of rate $\lambda$, as described in \eqref{eq:poisson} in the main text.
We consider a worst-case scenario and assume that Quicksort samples a comparison outcome for every pair of items.
Let $z_{ij}$ be the indicator random variable of the event ``the comparison between $i$ and $j$ resulted in an error''.
By Lemma~\ref{lem:dispbound}, we have
\begin{align}
\label{eq:probbound}
\Disp{\sigma} \le 2 \sum_{i < j}\; \Abs{i - j} z_{ij}
\end{align}
In the following, we will bound some of the statistical properties of the random variables $\{ z_{ij} \}$.
We start with a lemma that bounds their mean.

\begin{lemma}
\label{lem:expz}
For any $1 \le i < j \le n$,
\begin{align*}
\Exp{z_{ij}} \le \left( \frac{\lambda}{\lambda + 1} \right)^{j-i}.
\end{align*}
\end{lemma}
\begin{proof}
Let $d_{ij} = \theta_i - \theta_j$ be the (random) distance between items $i$ and $j$.
This distance is a sum of $k = j-i$ independent exponential random variables, and therefore $d_{ij} \sim \text{Gamma}(k, \lambda)$.
The comparison outcome is generated as per the BT model; conditioned on the distance $d_{ij}$, the random variable $z_{ij}$ is a Bernoulli trial with probability $[1 + \exp(d_{ij})]^{-1}$.
Therefore, we have that
\begin{align*}
\Exp{z_{ij}} \le \Exp{\exp(-d_{ij})} = \left( \frac{\lambda}{\lambda + 1} \right)^k
\end{align*}
\end{proof}

Next, we bound their covariance.
Note that the random variables $\{ z_{ij} \}$ are in general \emph{not} unconditionally independent.
They become independent only when conditioned on $\bm{\theta}$.

\begin{lemma}
\label{lem:covz}
For any $1 \le i < j \le n$ and any $1 \le u < v \le n$, let $A = \intint{i}{j\!-\!1}$ and $B = \intint{u}{v\!-\!1}$.
\begin{align*}
\Cov{z_{ij}}{z_{uv}} \le
\begin{dcases}
0
    & \text{if $A \cap B = \varnothing$,} \\
\left( \frac{\lambda}{\lambda + 1} \right)^{j - i}
    & \text{if $A = B$,} \\
\left( \frac{\lambda + 1}{\lambda + 2} \right)^{j - i + v - u}
    & \text{otherwise.}
\end{dcases}
\end{align*}
\end{lemma}
\begin{proof}
If $A$ and $B$ are disjoint, the distances $d_{ij}$ and $d_{uv}$ are independent random variables.
Conditioned on the distances, the comparison outcomes are independent Bernoulli trials, and we conclude that $z_{ij}$ and $z_{uv}$ are independent.
In the two remaining cases, we bound $\Exp{z_{ij}z_{uv}} \ge \Cov{z_{ij}}{z_{uv}}$.
If $A = B$, then $z_{ij} = z_{uv}$ and we have
\begin{align*}
\Exp{z_{ij} z_{uv}} = \Exp{z_{ij}^2} = \Exp{z_{ij}}
\end{align*}
and we apply Lemma~\ref{lem:expz}.
Finally, if $A$ and $B$ are neither equivalent nor disjoint, the two comparison outcomes are independent Bernoulli trials conditioned on the distances $d_{ij}$ and $d_{uv}$, but the distances are not independent.
Consider the case where $i < u < j < v$.
Even though $d_{ij}$ and $d_{uv}$ are dependent, the distances $d_{iu}$, $d_{uj}$, $d_{jv}$ are independent Gamma random variables of rate $\lambda$ and shape $u-i$, $j-u$ and $v-j$, respectively, and
\begin{align*}
\Exp{z_{ij}z_{uv}} &\le \Exp{\exp\{-(d_{iu} + d_{uj}) - (d_{uj} + d_{jv})\}} \\
    &= \left( \frac{\lambda}{\lambda + 1} \right)^{u-i} \left( \frac{\lambda}{\lambda + 2} \right)^{j-u} \left( \frac{\lambda}{\lambda + 1} \right)^{v-j}
    \le \left( \frac{\lambda + 1}{\lambda + 2} \right)^{j-i + v-u}
\end{align*}
The other cases are treated analogously.
\end{proof}

Lemmas~\ref{lem:expz} and~\ref{lem:covz} will be useful in proving the first part of Theorem~\ref{thm:quickdisp}.
For the second part, we need a result from \citet{ailon2008reconciling}, which characterizes the pairwise marginals of the distribution over rankings induced by Quicksort with comparisons sampled from a BT model.

\begin{theorem}[\citealp{ailon2008reconciling}, Theorem~$4.1$]
\label{thm:stability}
Let $\sigma$ be the output of Quicksort using comparison outcomes sampled from $\BT(\bm{\theta})$.
Then, for any $i, j \in [n]$,
\begin{align*}
\Prob{\sigma(i) < \sigma(j) \mid \bm{\theta}} = p(i \prec j \mid \bm{\theta})
\end{align*}
\end{theorem}

Note that the result is non-trivial as $i$ and $j$ might not have been directly compared to each other: their relative position might have been deduced by transitivity from other comparison outcomes.
We are now ready to prove Theorem~\ref{thm:quickdisp}.

\begin{proof}[Proof of Theorem~\ref{thm:quickdisp}]
We begin with the first part of the theorem, which bounds the displacement \Disp{\sigma}.
For clarity of exposition, we use the notation $z_{i \to k}$ instead of $z_{ij}$ if $j = i+k$.
Using~\eqref{eq:probbound} and Lemma~\ref{lem:expz}, we can bound the expected displacement as
\begin{align*}
\Exp{\Delta}
    \le \sum_{i=1}^{n-1} \sum_{k=1}^{n-i} 2k \Exp{z_{i \to k}}
    \le n \sum_{k = 1}^{\infty} 2k \left( \frac{\lambda}{\lambda + 1} \right)^{k} = 2n \lambda (\lambda + 1).
\end{align*}
In a similar way, using Lemma~\ref{lem:covz}, we can bound the variance of the displacement as
\begin{align*}
\Var{\Delta}
    &\le \sum_{i=1}^{n-1} \sum_{k=1}^{n-i} 4k^2 \Var{z_{i \to k}}
     + 2\sum_{i=1}^{n-1} \sum_{k=1}^{n-i} 2k
         \sum_{u=i+1}^{i+k} \sum_{\ell=1}^{n-u} 2\ell \Cov{z_{i \to k}}{z_{u \to \ell}} \\
    &\le n \sum_{k=1}^{\infty} 4k^2 \left( \frac{\lambda}{\lambda + 1} \right)^k
     + 2n \sum_{k=1}^{\infty} 2k^2 \left( \frac{\lambda + 1}{\lambda + 2} \right)^k
         \cdot \sum_{\ell=1}^{\infty} 2\ell \left( \frac{\lambda + 1}{\lambda + 2} \right)^\ell \\
    &\le 1500 n(\lambda^5 + 1).
\end{align*}
% Expressions used to solve the series on Wolfram Alpha:
% A: `sum from 1 to infinity of k^2 (x / (x+1))^k` -> x(x + 1)(2x + 1)
% B: `sum from 1 to infinity of k^2 ((x+1) / (x+2))^k` -> (x+1)(x+2)(2x+3)
% C: `sum from 1 to infinity of k ((x+1) / (x+2))^k` -> (x+1)(x+2)
% var(delta) < n*(4*A + 8*B*C)
Combining the bounds for the mean and the variance with Chebyshev's inequality, we have that
\begin{align*}
\Prob{\Disp{\sigma} \ge 50 n (\lambda^2 + 1)} \le \lambda / n,
\end{align*}
which concludes the proof of the first part of the claim.

The second part of the theorem bounds the maximum displacement for any single item.
We start by showing that with high probability, there is no pair of items separated by at least $\BigO{\lambda \log n}$ positions that is ``flipped'' in the output of Quicksort.
Let $i$ and $j$ be two items such that $i < j$ and let $k = \Abs{i - j}$.
Then $d_{ij} \sim \DGamma{k, \lambda}$, and using a Chernoff bound we obtain
\begin{align*}
\Prob{d_{ij} \le k/(e\lambda)} \le \exp(-k/e).
\end{align*}
If $k \ge 3 (\lambda + 1)e \log n$, we find that
\begin{align}
\label{eq:maxz1}
\Prob{d_{ij} \le k/(e\lambda) } \le \Prob{d_{ij} \le 3 \log n} \le n^{-3}.
\end{align}
Using the fact that the pairwise marginals of Quicksort match the pairwise comparison outcome probabilities (Theorem~\ref{thm:stability}), we find
\begin{align}
\label{eq:maxz2}
\Prob{\sigma(j) < \sigma(i)}
    = p(j \prec i)
    \le \exp(-3 \log n) = n^{-3}.
\end{align}
Combining \eqref{eq:maxz1} and \eqref{eq:maxz2}, and using a union bound over the $\tbinom{n}{k}$ pairs, we see that with probability $1 - 1/n$ there is no pairs of items $(i, j)$ separated by at least $3 (\lambda + 1)e \log n$ position with $i < j$ but $\sigma(j) < \sigma(i)$.
Finally, suppose that there is an $i$ such $\Abs{\sigma(i) - i} = k$.
Without loss of generality, we can assume that $i < \sigma(i)$.
This means that there are $k$ items larger than $i$ that are on the left of $i$ in $\sigma$.
In particular, there is an item $j > i$ such that $\Abs{i - j} \ge k$ and $\sigma(j) < \sigma(i)$.
This concludes the proof.
\end{proof}

%%%%%%%%%%%%%%%%%%%%%%%%%%%%%%%%%%%%%%%%%%%%%%%%%%%%%%%%%%%%%%%%%%%%%%%%%
\subsection{Theorem \ref{thm:multidisp}}
\label{sec:pfmdisp}

In order to prove Theorem~\ref{thm:multidisp}, we first need a basic result on the order statistics of exponential random variables.
Let $x_1, \ldots, x_n$, be i.i.d. exponential random variables of rate $\lambda$.
Let $x_{(1)}, \ldots, x_{(n)}$ be their order statistics, i.e., the random variables arranged in increasing order.
Then,
\begin{align}
\label{eq:expordstat}
x_{(i)} = \sum_{j = 1}^{i} \frac{1}{n - j + 1} y_j,
\end{align}
where $y_1, \ldots, y_n$ are i.i.d. exponential random variables of rate $\lambda$ \citep[see, e.g.,][Section 4.6]{arnold2008first}.

\begin{proof}[Proof of Theorem~\ref{thm:multidisp}]
We consider the order statistics of the $n - 1$ i.i.d. exponential random variables $x_1, \ldots, x_{n-1}$ which define the distances between neighboring items.
Let $\hat{n} = \ceil{n / \log^2 n}$, and denote by $B \subset [n]$ the set of items at both ends of $x_{(1)}, \ldots, x_{(\hat{n} - 1)}$.
These ``bad'' items are close to their nearest neighbor, and we simply invoke Theorem~\ref{thm:quickdisp} to claim that each of these items is shifted by at most \BigO{\lambda \log n} positions with high probability.
%Nevertheless, note that $B$ represents a vanishing fraction of the items.
Consider now the ``good'' items, i.e., those in $G = [n] \setminus B$.
Using \eqref{eq:expordstat} and for $n$ large enough,
\begin{align*}
\Prob{x_{(\hat{n})} \le 1/(e \lambda \log^2 n)}
    \le \Prob{\sum_{j=1}^{\hat{n}} y_j / n \le 1/(e \lambda \log^2 n)}
    \le \exp(-\hat{n} / e) \le 1 / n.
\end{align*}
The second-to-last inequality follows from a Chernoff bound similar to that used in the proof of Theorem~\ref{thm:quickdisp}.
Therefore, with high probability all items in $G$ are at distance larger than $c / (\lambda \log^2 n)$ from their nearest neighbor.

We will now show that after $m = \BigO{\lambda^2 \log^5 n}$ runs of Quicksort, $\hsigma(i) = i$ with high probability for all $i \in G$.
Let $i \in G$, $j \in [n]$ be a pair of items, and without loss of generality assume that $i < j$.
Let $t_k$ be the indicator random variable for the event ``$\sigma(i) < \sigma(j)$ in the $k$-th run of Quicksort'', and let $p = \Prob{t_k = 1}$.
Then, using Theorem~\ref{thm:stability},
%As $d_{ij} \ge 1 / (e \lambda \log^2 n)$ with high probability,
\begin{align*}
p - \frac{1}{2}
    = p(i \prec j) - \frac{1}{2} = \frac{1 - \exp(-d_{ij})}{2[1 + \exp(-d_{ij})]}
    \ge \frac{1 - \exp[- 1/(e \lambda \log^2 n)]}{4}
    \ge \frac{1}{8e \lambda \log^2 n}
\end{align*}
with high probability.
In the last inequality, we used the fact that $1 - e^{-x} \ge x/2$ for $x \in [0, 1]$.
The random variables $t_1, \ldots, t_m$ are independent Bernoulli trials, and using a Chernoff bound we obtain
\begin{align*}
\Prob{\hsigma(j) < \hsigma(i)} = \Prob{ \sum_{k = 1}^m t_k \le n/2}
    \le \exp[-2m(p - 1/2)^2] \le \exp \left[ -\frac{m}{32e^2 \lambda^2 \log^4 n} \right].
\end{align*}
With $m = 96e^2 \lambda^2 \log^5 n$, we have $\Prob{\hsigma(j) < \hsigma(i)} \le n^{-3}$, and using a union bound we see that with probability $1 - 1/n$ we have $\hsigma(i) = i$ for all $i \in G$.
Therefore, the total displacement is
\begin{align*}
\Disp{\hsigma} = \sum_{i \in B} \Abs{\hsigma(i) - i}
    \le \Abs{B} \cdot 3(\lambda + 1) e \log n
    = \BigO{\lambda n / \log n}.
\end{align*}
This concludes the proof.
\end{proof}

%%%%%%%%%%%%%%%%%%%%%%%%%%%%%%%%%%%%%%%%%%%%%%%%%%%%%%%%%%%%%%%%%%%%%%%%%
\section{Discriminating the Closest Items}  %%%%%%%%%%%%%%%%%%%%%%%%%%%%%
\label{app:adjacent}

The distance between the two closest items is $d_{\min} = \min_i \Abs{\theta_{i+i} - \theta{i}} = \min_i x_i$, i.e., the minimum of $n-1$ independent exponential random variables of rate $\lambda$.
Therefore, $d_{\min} \sim \DExp{(n-1)\lambda}$, and for $n \ge 2$ with probability at least $1 - e^{-1/2} \approx 0.39$ we have $d_{\min} \le (\lambda n)^{-1}$.
Suppose that we compare the two closest items $m$ times, and let $z_i$ be the indicator random variable for the event ``the outcome of the $i$-th comparison is incorrect''.
Assuming that $d_{\min} \le (\lambda n)^{-1}$ and that $\lambda n \ge 1/2$,
\begin{align*}
\Prob{z_i = 0}
    \le \frac{1}{1 + \exp[-1 / (\lambda n)]} \le \frac{1}{2 - 1/(\lambda n)}
    = \frac{1}{2} \cdot \left( 1 + \frac{1}{2\lambda n - 1} \right)
    \le \frac{1}{2} \exp \left[ \frac{1}{2\lambda n - 1} \right],
\end{align*}
where we used the inequality $e^{x} \ge 1 + x$ twice.
Given the $m$ comparison outcomes, we use a majority vote to decide the relative order of the two items.
The probability of making the correct decision is
%Based on these $m$ outcomes, we use a majority vote to decide the respective order of the items.
\begin{align*}
\Prob{\sum_{i = 1}^m z_i \le m/2}
    &\le \sum_{k = 1}^{m/2} \binom{m}{k} \Prob{z_i = 0}^m
     \le \exp \left[ \frac{m}{2\lambda n - 1} \right] \cdot 2^{-m} \sum_{k = 1}^{m/2} \binom{m}{k} \\
    &= \frac{1}{2} \exp \left[ \frac{m}{2\lambda n - 1} \right].
\end{align*}
Therefore, if $m = \LittleO{\lambda n}$ the probability of making a mistake is bounded from below by a positive constant.

%%%%%%%%%%%%%%%%%%%%%%%%%%%%%%%%%%%%%%%%%%%%%%%%%%%%%%%%%%%%%%%%%%%%%%%%%
\section{Additional Figures}  %%%%%%%%%%%%%%%%%%%%%%%%%%%%%%%%%%%%%%%%%%%
\label{app:figures}

In this section, we present a few additional figures that complement the ones presented in Section~\ref{sec:experiments} of the main text.

\begin{figure}[t]
\centering
\includegraphics[width=0.67\linewidth]{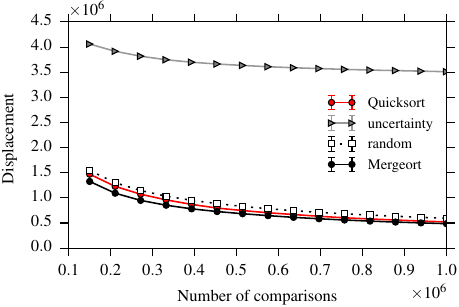}
\caption{
Results on the GIFGIF dataset.
The experiment is repeated \num{10} times, and we report the mean and the standard deviation.
The variant of uncertainty sampling performs extremely poorly.
}
\label{fig:gifgif}
\end{figure}

Figure~\ref{fig:gifgif} presents the results on the GIFGIF dataset including a variant of uncertainty sampling.
This variant samples, at each iteration, $n-1$ comparisons consisting of adjacent pairs in the ranking $\hat{\bm{\theta}}$.
This strategy performs surprisingly poorly.

\begin{figure*}[t]
\centering
\includegraphics[width=\linewidth]{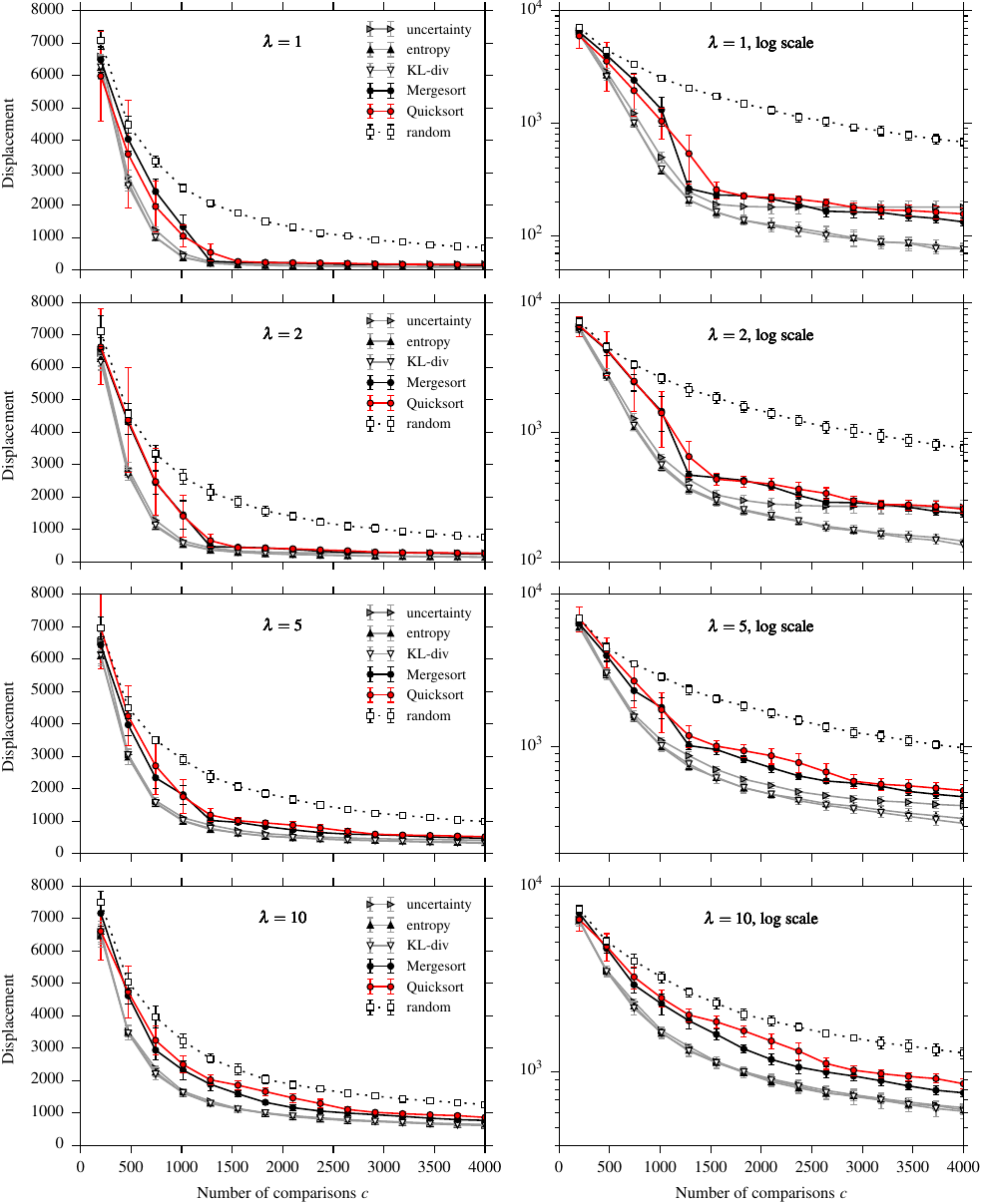}
\caption{
Results on synthetic datasets for $n = 200$ and increasing values of $\lambda$.
Every experiment is repeated \num{10} times, and we report the mean and the standard deviation.
}
\label{fig:baselines2}
\end{figure*}

Figure~\ref{fig:baselines2} presents results on synthetic datasets with $n = 200$ and $\lambda \in \{ 1, 2, 5, 10 \}$.
For the reader's convenience, we plot every graph on both a linear and a logarithmic scale.
Unsurprisingly, the gains of adaptive sampling are greater when the noise is smaller.

\hyphenation{wahrscheinlichkeits-rechnung} % :-)
\bibliography{robustsort}

\end{document}